\documentclass[letterpaper, 10 pt, journal, twoside]{IEEEtran}
\usepackage{graphicx}
\usepackage{amsmath}
\usepackage{amssymb}
\usepackage{makecell}
\usepackage{multirow}
\usepackage{multicol}
\usepackage{caption}

\ifCLASSINFOpdf
\else
\fi
\begin{document}
%
\title{Recursive Cross-View: Use Only 2D Detectors to Achieve 3D Object Detection without 3D Annotations}
%
%
%

\author{Shun Gui$^{1}$, and Yan Luximon$^{1}$
\thanks{Manuscript received: May, 25, 2023; Revised August, 10, 2023; Accepted August, 10, 2023.}
\thanks{This paper was recommended for publication by Editor Cadena Lerma, Cesar  upon evaluation of the Associate Editor and Reviewers' comments.
This work was supported by the Laboratory for Artificial Intelligence in Design (Project Code: RP1-3), Innovation and Technology Fund, Hong Kong Special Administrative Region} 
\thanks{$^{1}$The authors are with the School of Design, The Hong Kong Polytechnical University, Hong Kong.
        {\tt\small shun.gui@connect.polyu.hk; yan.luximon@polyu.edu.hk}}%
\thanks{}%
\thanks{Digital Object Identifier (DOI): see top of this page.}
}
%
%

\markboth{IEEE Robotics and Automation Letters. Preprint Version. Accepted August, 2023}
{Gui \MakeLowercase{\textit{et al.}}: Recursive Cross-View} 

%



\maketitle

\begin{abstract}
Heavily relying on 3D annotations limits the real-world application of 3D object detection. In this paper, we propose a method that does not demand any 3D annotation, while being able to predict fully oriented 3D bounding boxes. Our method, called Recursive Cross-View (RCV), utilizes the three-view principle to convert 3D detection into multiple 2D detection tasks, requiring only a subset of 2D labels. We propose a recursive paradigm, in which instance segmentation and 3D bounding box generation by Cross-View are implemented recursively until convergence. Specifically, our proposed method involves the use of a frustum for each 2D bounding box, which is then followed by the recursive paradigm that ultimately generates a fully oriented 3D box, along with its corresponding class and score. Note that, class and score are given by the 2D detector. Estimated on the SUN RGB-D and KITTI datasets, our method outperforms existing image-based approaches. To justify that our method can be quickly used to new tasks, we implement it on two real-world scenarios, namely 3D human detection and 3D hand detection. As a result, two new 3D annotated datasets are obtained, which means that RCV can be viewed as a (semi-) automatic 3D annotator. Furthermore, we deploy RCV on a depth sensor, which achieves detection at 7 fps on a live RGB-D stream. RCV is the first 3D detection method that yields fully oriented 3D boxes without consuming 3D labels.
\end{abstract}


%
\IEEEpeerreviewmaketitle

\section{INTRODUCTION}

\begin{figure}[t]
\begin{center}
   \includegraphics[width=0.8\linewidth]{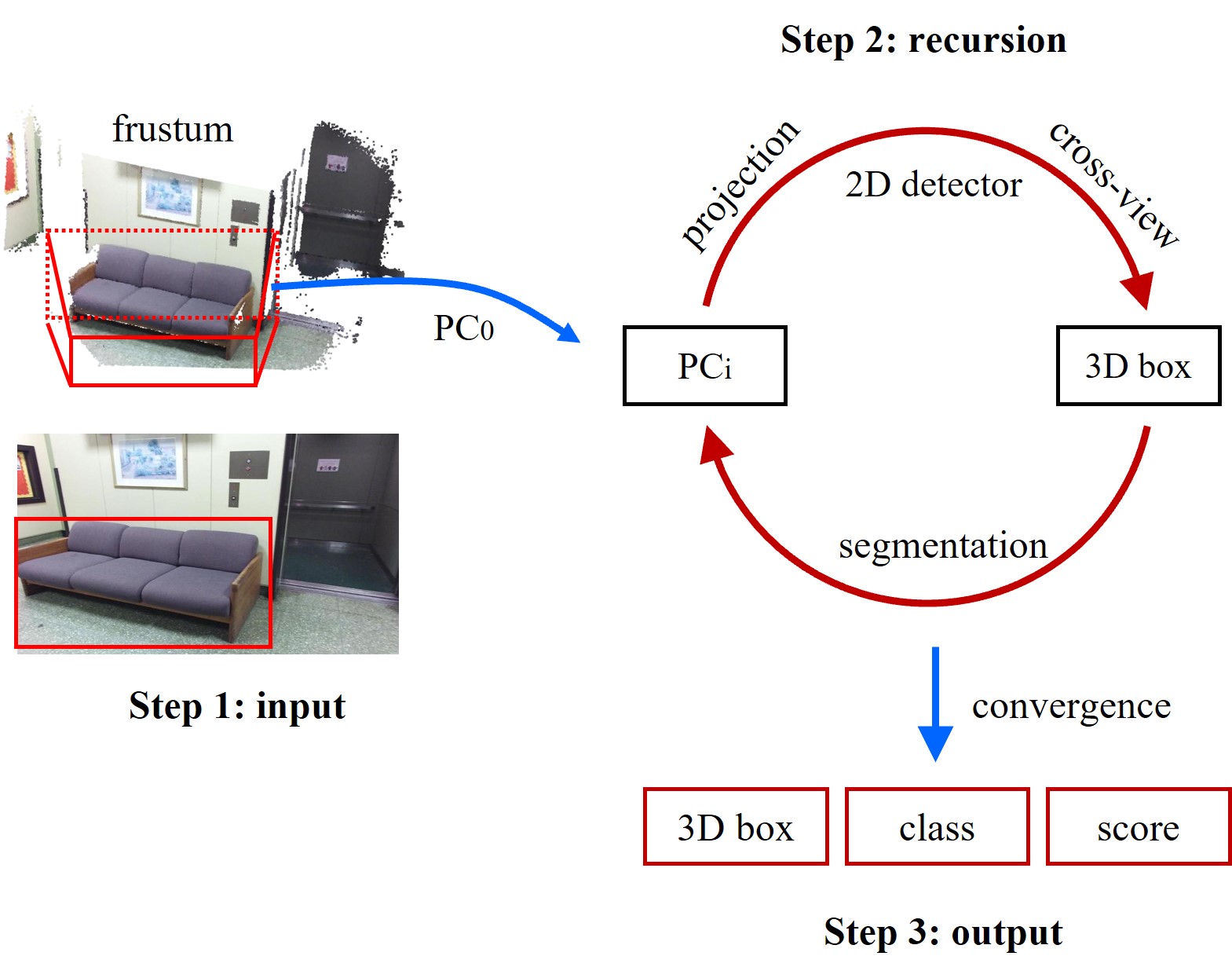}
\end{center}
   \caption{{\bf Overview of RCV.} Step 1: execute 2D detector on an image and propose frustums on the point cloud. Step 2: perform recursion. Step 3: output. Note that, class and score are given by 2D detector. See Fig. 4 for more details on the recursion.}
\label{fig:long}
\label{fig:onecol}
\end{figure}

3D object detection aims to locate, classify, and generate 3D bounding boxes of objects in a scene. With advancements in 3D sensors, annotated datasets, and deep learning techniques, 3D detection has made remarkable progress in recent years, and it could play a crucial role in various applications such as autonomous driving, robot navigation, robot manipulation, and human-robot interaction. Existing methods can be broadly classified into several
categories, namely image-based [4,9,10,32-34], projection-based [1,17], voxel-based
[12,14,18], and point-based [5,13,15,16,18,19]. However, most existing works focus on autonomous driving and indoor object detection tasks, thanks to publicly available datasets such as KITTI [24] and SUN RGB-D [25]. Generally, 3D object detection involves training a model to perform regression on a set of 3D bounding boxes that serve as ground truths. Nevertheless, what if one needs to detect a new object in a new scene?

A common solution is to first label enough 3D annotations and then train a network. However, manual 3D labeling on data collected by RGB-D sensors or LiDAR is both laborious and expensive [20,21]. For example, the authors of SUN RGB-D hired and trained 18 oDesk workers who spent a total of 2051 hours annotating data [25]. Obviously, this approach is not very efficient for achieving fast (within several hours) real-world applications. Many studies adopted a fully supervised learning paradigm, that is, representing data, learning features, and classifying and regressing. However, the efficacy and reliability heavily rely on the availability of precise 3D annotations. To alleviate the need for 3D annotations, weakly supervised [20,21,42], semi supervised [8,22], and self-supervised [23] strategies on 3D object detection were proposed. [20] proposed a weakly supervised framework that leveraged BEV center-click annotations and several hundred 3D labels to train a model. However, it still consumes 3D annotations. So, can we just rely on other easily available labels to achieve 3D detection? 

While certain 3D detection approaches have demonstrated strong performance in certain scenarios, their effectiveness is largely constrained to detecting objects with a vertical orientation [1-3,5-7,13-16]. Specifically, the generated boxes are restricted to a vertical orientation and do not account for any potential roll or pitch, thus limiting the applicability of these methods. If one wants to predict fully oriented boxes, he/she has to enlarge the output dimension, inevitably increasing the detection difficulty. Also, almost no dataset is available. Fully oriented detection has a wider application scope, for example robot manipulation and human-robot interaction. If a 3D detection method that can detect fully oriented objects can be quickly implemented on novel 3D sensors in various scenarios, it would be highly beneficial for real-world applications.

In this work, we propose a simple yet effective 3D detection method named RCV that does not use 3D annotations. Fig. 1 demonstrates the overview of RCV. By exploring the principle of three-view drawings, we convert 3D detection into several 2D detection tasks. Following a recursive paradigm, RCV can achieve instance segmentation and predict fully oriented 3D bounding boxes. RCV offers several advantages over existing works. First, it does not rely on 3D annotations, which enhances its applicability in the real world. Due to mature 2D detectors, RCV only requires some 2D labels to achieve 3D detection and can inherit the properties of 2D detectors, such as robustness and generalization. Second, RCV can predict fully oriented boxes, which greatly extends the range of application. Third, RCV can be quickly deployed to new 3D sensors in new real-world scenes. Moreover, once trained, RCV can be viewed as a 3D annotation tool to simplify manual labeling or formulate datasets for pretraining.

Estimated on the SUN RGB-D and KITTI datasets, our method outperforms existing image-based approaches. Particularly, our method is extremely data efficient, and it outperforms existing image-based methods dramatically in the 3D detection of Pedestrian and Cyclist in KITTI using only 25\% of the training data. In the real-world experiments, RCV can achieve 3D detection by training only on 2D images and 2D labels. Once trained, RCV can be used as a (semi-) automatic 3D annotator, and two datasets are formulated via RCV. Our approach provides a solution that uses some 2D annotations to achieve 3D detection, which is an important practical contribution of this work.

Our contributions are as follows:

\begin{itemize}
\item We propose a new method for fully oriented 3D object detection without 3D annotations. Also, our method achieves state-of-the-art performance on SUN RGB-D and KITTI datasets.
\item We formulate a (semi-) automatic 3D annotator, which can be used to label 3D bounding boxes in new scenarios.
\item We deploy our method to a real 3D sensor, achieving real-time detection of a new object in a new scene.
\end{itemize}

In the following sections, Section II outlines the core components of RCV. Section III presents experiments conducted on open-source datasets as well as our own collected data. Discussion and future work are presented in Section IV. Finally, the conclusion is presented in Section V.

\section{Recursive Cross-View}

\subsection{Conversions between 3D and 2D}

Since manually annotating 3D bounding boxes is quite labor-intensive and expensive, it motivates us to consider how to circumvent this problem so that 3D detection algorithms can be quickly implemented into previously unseen scenarios/objects. RCV is inspired by the principle of engineering drawing, in which a 3D object can be fully depicted by three views (the left-top subimage in Fig. 2) and vice versa (the right-top subimage in Fig. 2). This correlation means that a 3D object can be restored using only three 2D views, this is the essential idea of RCV. However, how to derive a 3D bounding box in this process? In 3D object detections, the aim is to produce a 3D bounding box for each object instead of reconstructing all of its details. A 3D bounding box can be derived by detecting the size and location of the 2D bounding boxes and then following the mechanism of three views, as shown in the left-bottom subimage in Fig. 2. During this process, 3D annotations are not required. Another benefit of RCV is that it can directly detect fully oriented bounding boxes without raising the difficulty of detection because it does not rely on regression. To justify this advantage, we formulate a “3D\_HAND” dataset, which includes annotations of fully oriented boxes, see Section III for more details.
\begin{figure}[t]
\begin{center}
   \includegraphics[width=0.7\linewidth]{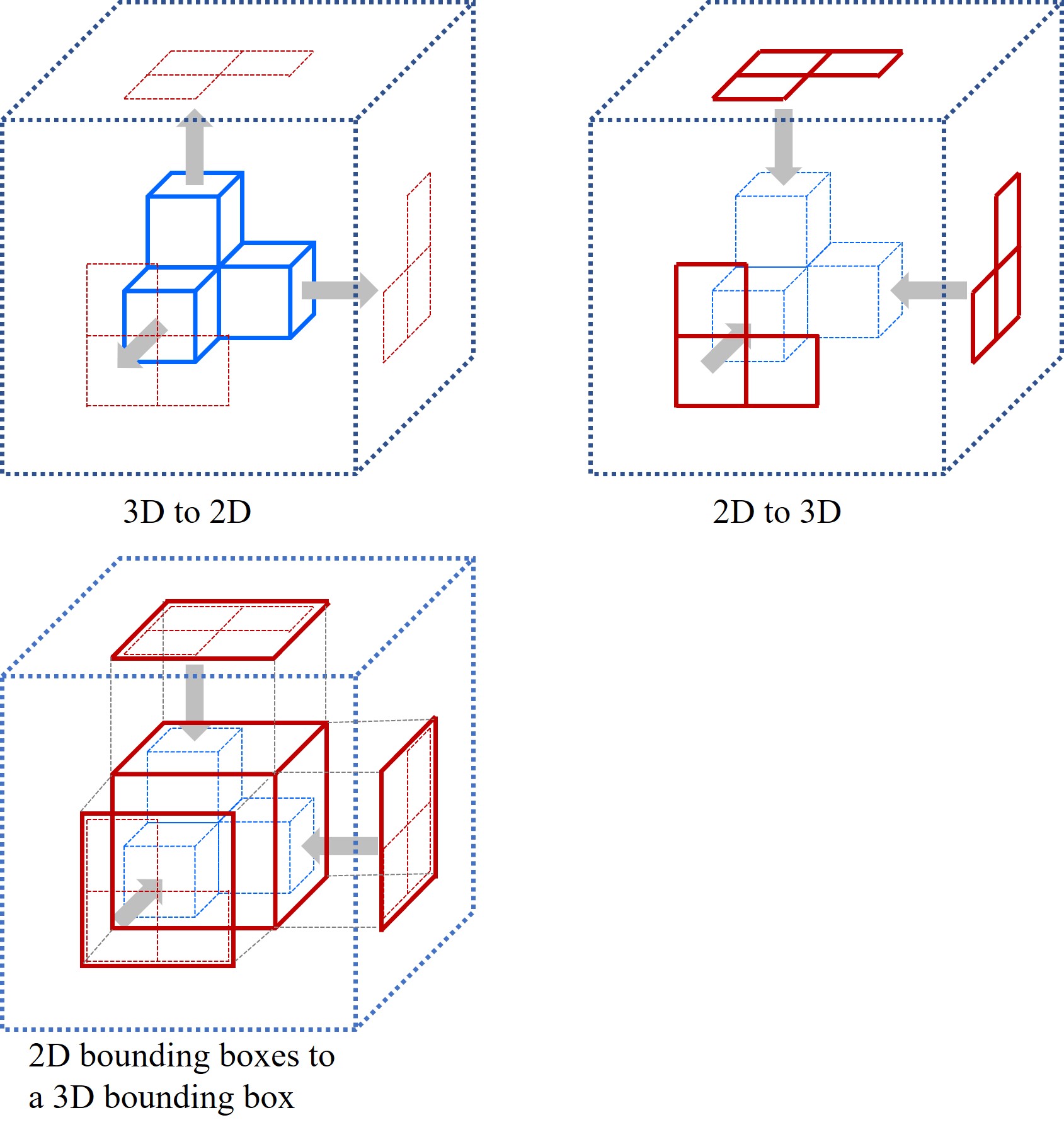}
\end{center}
   \caption{{\bf Conversion between 3D and 2D.} The left-top and right-top subimages are three views, and the left-bottom subimage is the derivation of 3D bounding box from three views.}
\label{fig:long}
\label{fig:onecol}
\end{figure}

\subsection{Perspective View}
RCV aims to gradually eliminate points that do not belong to the object, while preserving volumetric regions that are occupied by the object but not rendered in the depth camera. Finally, a bounding box is generated for each object. With this guideline in mind, we formulate each step of RCV. The first step called “Perspective View”, involves processing the raw data, i.e., RGB images and point clouds captured by the depth camera.  This step is similar to the operation used by F-PointNets [3], in which a frustum can be derived according to the projection matrix of depth camera and the 2D bounding box. Fig. 3 illustrates this process. RCV, however, leverages totally different ideas with F-PointNets apart from this operation.

The reasons why we leverage perspective view first include (1) it is straightforward to capture RGB-images from cameras; (2) the perspective view is larger than the orthographic view, which is helpful for handling detections in large scene, such as self-driving tasks. After this step, a very coarse 3D bounding box enclosing the point cloud in the frustum can be generated, but it is not mandatory. This depends on the specific scenarios, for example, we generate bounding boxes at the first step on SUN RGB-D benchmark since the presence of significant object occlusion and background clutter. However, coarse 3D boxes are not generated at the first step in the experiment conducted on KITTI. Note that RCV exploits YOLOv5 as the 2D detector at all steps.

\begin{figure}[t]
\begin{center}
   \includegraphics[width=0.8\linewidth]{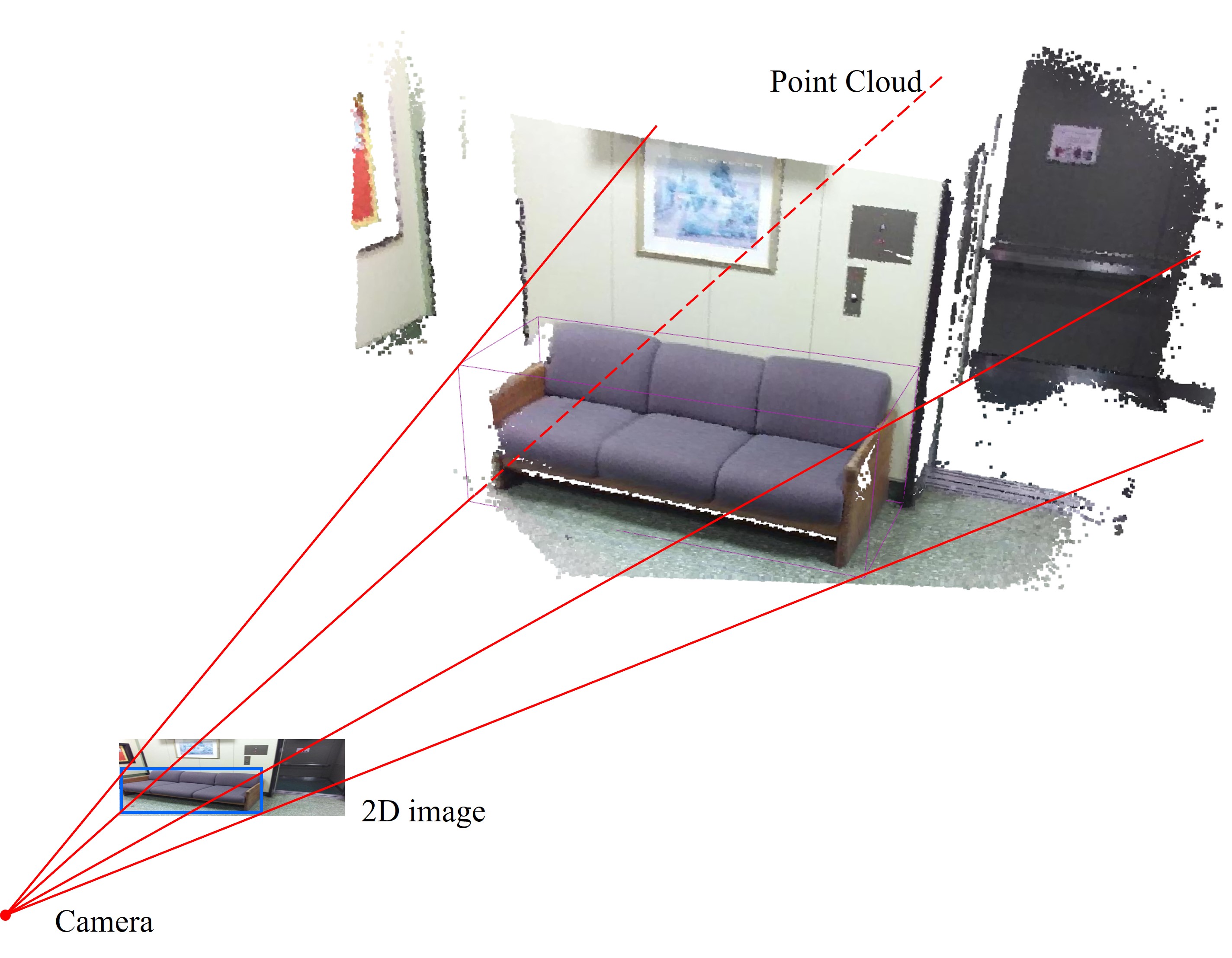}
\end{center}
   \caption{{\bf Perspective view.} A 2D bounding box can be obtained from 2D detector, then a frustum can be derived.}
\label{fig:long}
\label{fig:onecol}
\end{figure}

\begin{figure}[t]
\begin{center}
   \includegraphics[width=0.8\linewidth]{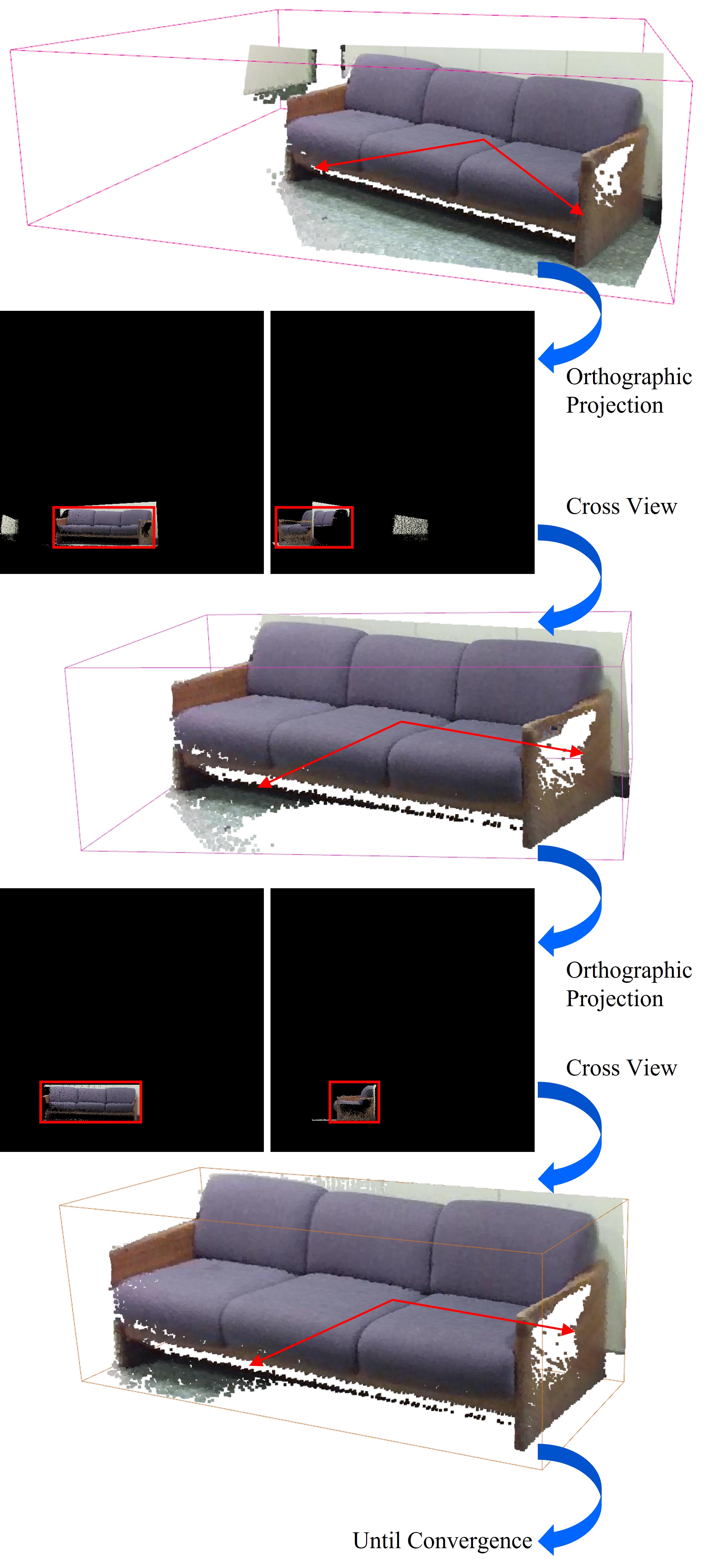}
\end{center}
   \caption{{\bf Recursively Cross-View.} Red arrows indicate orthographic view direction, blue curved arrows indicate projection and Cross-View that generates a 3D bounding box.}
\label{fig:long}
\label{fig:onecol}
\end{figure}

\subsection{Recursive Orthographic Cross-View}
For each frustum obtained through perspective view, we leverage the idea of divide and conquer to detect objects in parallel. We then apply the orthographic Cross-View recursively to generate the corresponding bounding box for each object based on the principle presented in the left-bottom subimage in Fig. 2. Any two views of the three views can be used to generate a deterministic box. Therefore, we choose front-view and side-view in all experiments. Fig. 4 demonstrates the recursive flow. The point cloud presented in the topmost portion of Fig. 4 is obtained from the frustum introduced by a 2D box. Next, we project these points along orthogonal axes to generate two RGB images, as shown in the second row of Fig. 4, with red arrows indicating the projection axes. Thereafter, we use YOLOv5 to detect objects on these two images, as shown in the two images in the second row of Fig. 4. As a result, the points that are not deemed to be objects are eliminated, as well as a more refined box can be obtained by performing the Cross-View, see the point cloud in the middle of Fig. 4. Recursively, we perform these operations on the retained point cloud, i.e. (1) computing the projection axes, (2) projecting RGB images for both views, (3) performing 2D detection to remove the external points, and (4) obtaining a new box through Cross-View, until convergence.

{\bf Convergence Conditions  } Several conditions are set to terminate the recursion. One of conditions is projection axes, that is, the red arrows shown in Fig. 4. We can see that the directions of the axes converge as the recursion proceed. The recursion terminates when their variation is less than a certain threshold. The second condition is the variation of the 3D bounding box, i.e., the recursion terminates when the difference between two consecutive boxes is less than a certain threshold. The third condition is to empirically fix the number of steps for recursion. The method for calculating the axes is presented in the next section.

{\bf Pseudo-view Images } The point cloud acquired by the depth camera contains spatial (XYZ) and color (RGB) data. The orthographic projection derives the corresponding pixel for each point based on the spatial data and preserves the color information. Finally, the projected images are obtained. We call these projected images 'pseudo-view' images.

{\bf Multi-object Detection in One Frustum  } The presence of occlusion can result in the existence of multiple objects within a frustum. Therefore, RCV is set to detect multiple objects on the pseudo-view images generated from the frustum's point cloud. Note that, only detections that have the same label as the 2D bounding box of the original image that proposes the frustum are preserved. For example, the topmost point cloud in Fig. 4 is from the frustum proposed by the 2D box in Fig. 3, and the label is “sofa”. Then we only detect multiple boxes with the label of “sofa” on the projected images in the second row of Fig. 4. After that, one or more 3D boxes are generated through Cross-View, followed by a recursive process for each, see Fig. 4. On the contrary, only one box is remained during the subsequent detections, which is used to refine the related 3D bounding box.

{\bf 3D Bounding Boxes  } For each point cloud in Fig. 4, we calculate the projection axis, and then obtain the next set of the point cloud and the box leveraging the Cross-View. Therefore, the transformation matrix ($T_n^{n+1}$) of two sets of point clouds and boxes of adjacent steps can be obtained from the coordinate system determined by the projection axes. Specifically, the middle point cloud in Fig. 4 is projected along the projection axis (red arrows), and the bottom point cloud and box are obtained after performing Cross-View. Inversely, the bottom point cloud and box can be transformed into the coordinate system of the middle point cloud according to the transformation matrix determined by the previous projection axis. Consequently, we can transform the last box into the original point cloud system through Eq. (1):
\begin{equation}
B_o = \prod_{i=0}^{N-1}T_i^{i+1}{B_N} 
\end{equation}
where $B_N$, 4 by 8 matrix, is the box generated at the Nth step of the recursion. $B_o$, 4 by 8 matrix, is the box corresponding to the original point cloud system. $T_i^{i+1}$ is a homogeneous transformation matrix with 4 by 4. Finally, we utilize non-maximum suppression (NMS) algorithm for all detected 3D bounding boxes, filtering the redundant detections.

\subsection{Projection Axes}
{\bf Camera Coordinate Axes  } It is straightforward to leverage camera axes as the projection directions. These axes adopted in all point clouds extracted from frustum, which is the first step of recursion, see the top point cloud in Fig. 4. The reason is that the point cloud in the frustum has a high probability of severe occlusion and cluttered background. The projection and Cross-View without any transformation can quickly perform preliminary detection.

{\bf Eigenvectors} The eigenvectors of point cloud are able to roughly depict its orientation, which thus can be used as projection axes. However, there is a limitation that they cannot represent the orientation of an object if only a small portion of the point cloud is acquired.

{\bf Normal Vectors  } The normal vectors can represent the orientation of the object, even when only a partial view of the object is available. We use normal vectors as the projection axes in our experiments. Specifically, we leverage K-Means algorithm to obtain the major normal vector of the point cloud. RCV could generate fully oriented 3D boxes, see Section III.D for more details.

\section{EXPERIMENTS}

We implement four experiments, namely (1) 3D detection performance on SUN RGB-D, (2) data efficiency on KITTI, (3) 3D annotator on our own data, and (4) real-time detection on a depth camera. As stated above, we do not formulate a new neural network. Thus, we only need to rely on the projected images to train YOLOv5, which is a simple and straightforward task. Note that we use default hyperparameters for all experiments.

\subsection{Obtaining 2D Annotations and Training data}
In this section, we present the method for obtaining 2D bounding boxes on both open datasets and our own datasets. For open datasets, due to 3D bounding boxes have already been annotated, the 2D bounding boxes can be obtained via the orthogonal projection of 3D bounding boxes	  , as shown in Fig. 5. Then the obtained images and 2D bounding boxes are utilized to train a 2D detector, which is utilized to formulate a RCV model. Although 3D bounding boxes are leveraged to derive 2D bounding boxes, they are not directly involved in model training. Next, we will introduce how to label 2D bounding boxes without using any 3D bounding boxes on our own dataset.

Manually annotating 3D bounding boxes is very challenging. Here, we demonstrate how to label 2D annotations on various scenarios or tasks for our method. Indoor 3D human dataset is used as an example. First, an image and a point cloud are captured by a depth camera, then we label 2D bounding boxes on the image, see the first row in Fig. 6. Next, the points in the frustum are retained, which is then used for projection to generate 2D images, see the second row in Fig. 6. The similar steps are repeated to generate two more 2D images, see the last row in Fig. 6. Particularly, we call these 2D images ‘pseudo-view’ images. 2D bounding boxes can be easily annotated on these images, which are then used to train a 2D detector. Finally, we can obtain a RCV model enabling detect 3D humans. In the Section III.D, we label 1,600 2D bounding boxes for 3D human detection and 530 2D bounding boxes for fully orientated 3D hand detection. A more detailed introduction to the 2D annotation can be accessed in the supplementary material.

Note that there are fundamental differences between our 2D annotation strategy and common 3D annotation strategy. While it is true that two 2D boxes from orthogonal views can form a 3D box, it does not necessarily guarantee the quality of the resulting 3D box since the correct projection axes are not known. In Fig. 4, the middle 3D box is not good enough to be used as a 3D annotation. 2D annotations are only used to enable 2D detector to detect objects in the projection images, regardless of the projection direction. We utilize a recursive process
to iteratively refine the 3D bounding box until convergence, resulting in a ’good’ 3D box. At each step, points outside the box are removed, and new projection axes - namely, the orientation of the 3D box - are obtained. Only the projection images at the first two steps in the ’Recursion’ are labelled, which cannot generate a good 3D annotation. Similarly, 2D boxes are derived from annotated 3D boxes in SUN RGB-D and KITTI, but our method still do not obtain the correct orientation of the object. Therefore, our method does not directly leverage 3D annotations in these open datasets.
\begin{figure}[t]
\begin{center}
   \includegraphics[width=0.9\linewidth]{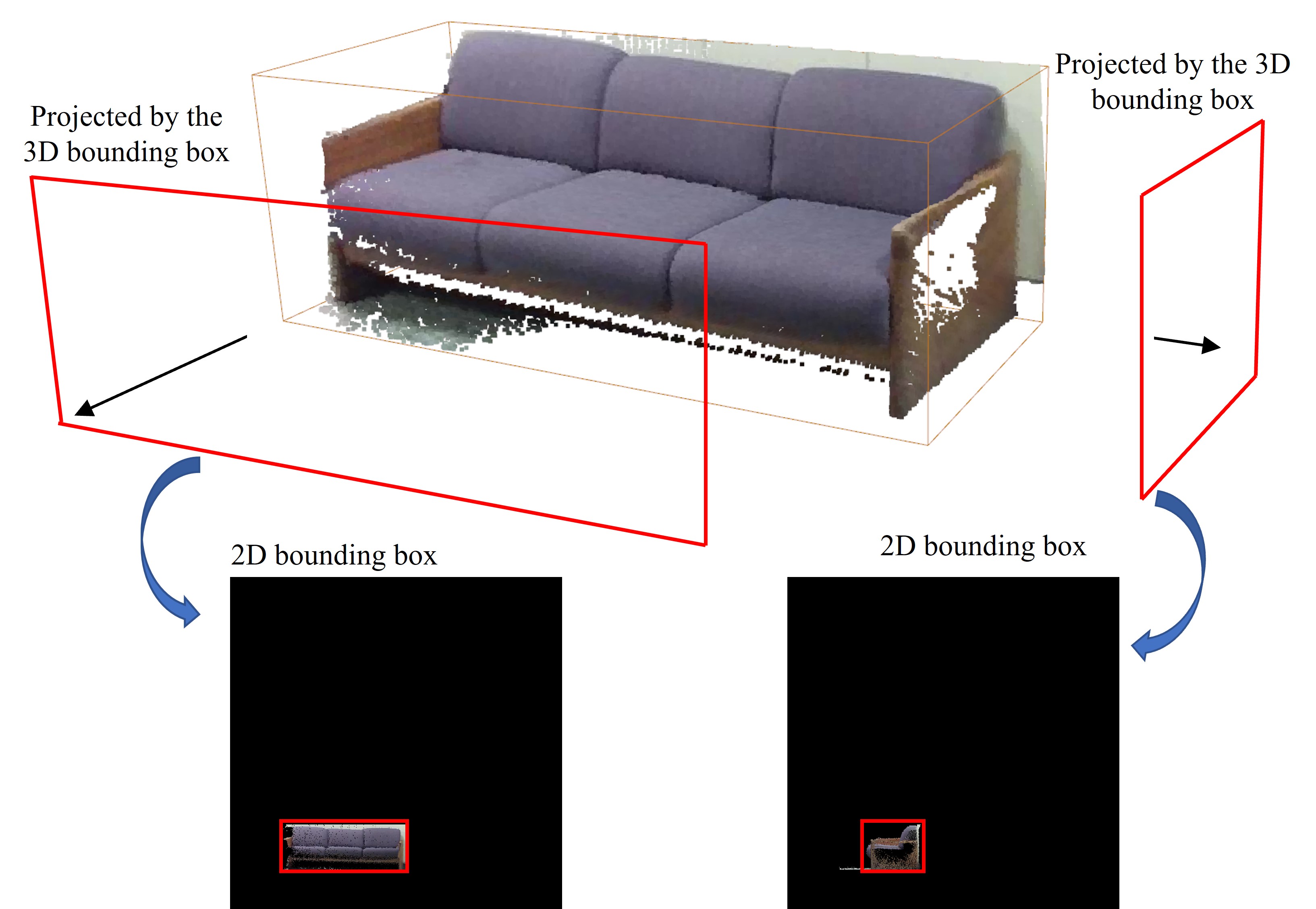}
\end{center}
   \caption{{\bf 2D bounding boxes projected by the 3D bounding box for SUN-RGBD dataset.}}
\label{fig:long}
\label{fig:onecol}
\end{figure}
\begin{figure}[t]
\begin{center}
   \includegraphics[width=0.8\linewidth]{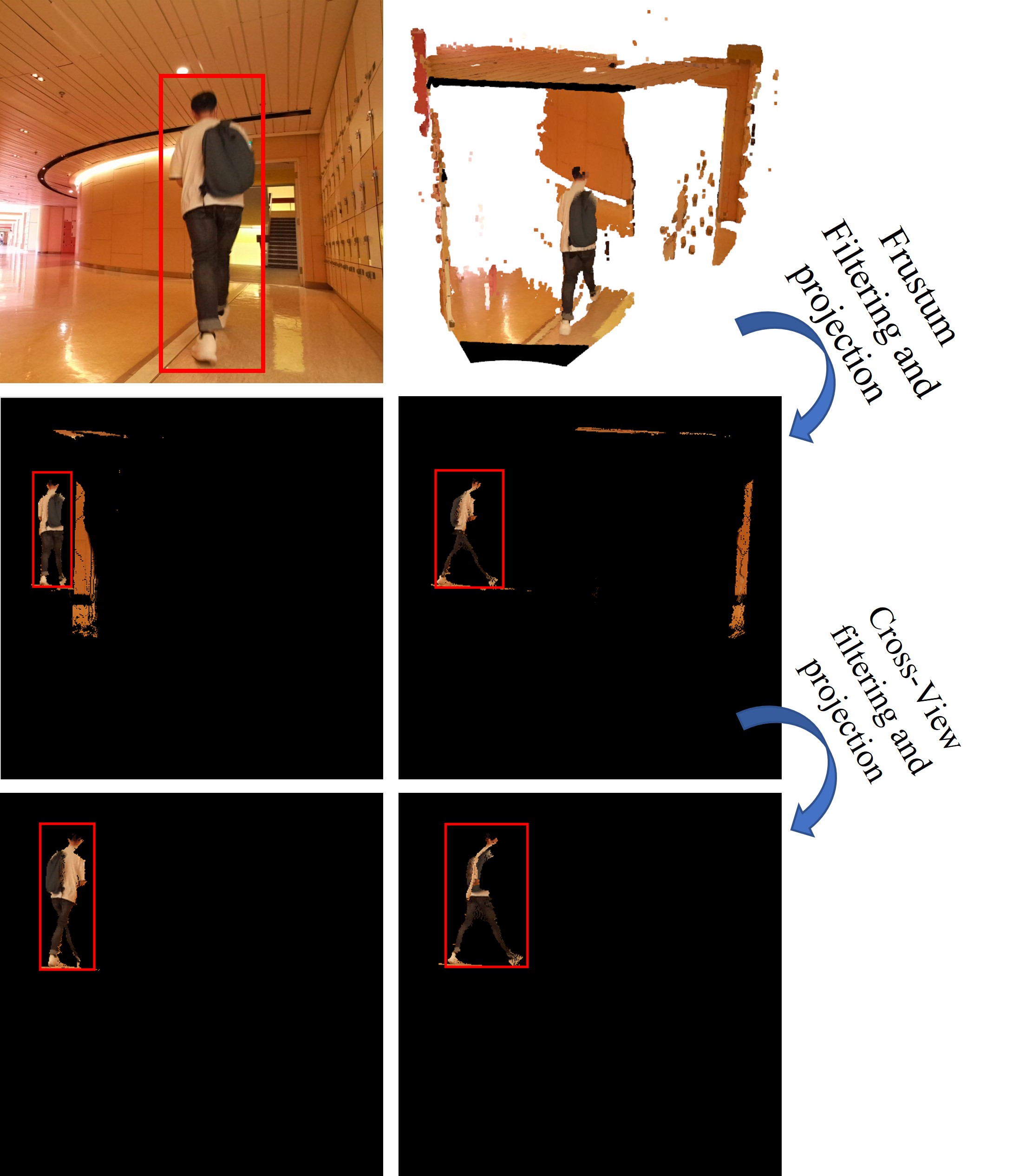}
\end{center}
   \caption{{\bf Manually 2D bounding box labeling method on our own dataset.}}
\label{fig:long}
\label{fig:onecol}
\end{figure}

\subsection{SUN RGB-D}
SUN RGB-D is an indoor 3D dataset with 5285 training samples and 5050 testing samples. We conduct a comparative experiment involving monocular 3D detection methods on this dataset. Our method takes several images as input, including a raw RGB image and several 'pseudo-view' images obtained through point cloud projection. Therefore, RCV has similar settings to some image-based 3D object detection methods. Monocular detection methods have been developed to detect 3D objects by combining a monocular image and geometric features or 3D world priori. In addition to this, some methods integrated a monocular image with depth maps for more accurate 3D object detection. LPCG-Monoflex [40] leveraged an image and LiDAR points to accomplish monocular 3D object detection in autonomous driving.

To compare RCV with monocular detection methods, we evaluate RCV on SUN-RGBD for 10 out of 37 object categories [30]. First, we convert all 3D objects into 2D images and 2D bounding boxes following the method in Section III.A, generating more than 100k images for training and more than 70k images for validation. Tab. I presents more details of training 2D detectors. In our experiment, we observe that the raw RGB image and ‘pseudo-view’ images exhibit considerable differences. Therefore, we train two separate 2D detectors: one for RGB images and one for 'pseudo-view' images. We adopt this setting for all experiments. Once the 2D detectors are trained, we can formulate RCV and leverage it to detect 3D bounding boxes on the validation set of SUN RGB-D. Tab. II demonstrates that RCV surpasses all previous methods and achieve state-of-the-art performance on this benchmark, without directly utilizing 3D annotations. Note that IM3D utilized extra data to train the model. Therefore, we do not compare our method with it.

\begin{table}
\begin{center}
\caption{{\bf Settings of training YOLO for 10 out of 37 object categories [30] in SUN-RGBD.}   The first row is the setting of the first step detection model, and the second row is the setting of the recursive detection model.}

\begin{tabular}{|l|c c c c c|}
\hline
& Model & Train no. & Val no. & Size & Device\\
\hline\hline
1 & YOLOv5x6 & 27,044 & 5050 &/ &3090\\
2 & YOLOv5x6 & 70,924 & 69848&640&3090\\
\hline
\end{tabular}
\end{center}

\end{table}

\begin{table*}
\begin{center}
\caption{{\bf 3D detection performance on SUN-RGB-D val. set for 10 out of 37 object categories [30]. }   The metric is average precision with 3D IoU threshold 0.15. We compare our scores with previous state-of-the-art monocular detection method. {\bf Bold} is used to highlight the best results. * means the method (IM3D) utilized extra data to train the model.}
\begin{tabular}{|l| p{1cm} c c p{0.5cm} p{0.5cm} p{0.5cm} p{0.5cm} p{0.7cm} p{0.7cm} p{0.7cm} p{0.7cm} p{0.7cm} p{0.7cm} | p{0.9cm}|}
\hline
 Method& Ref. & input & label & sink & bed & lamp & chair & desk & dresser & ntstand & sofa & table & cab & mAP\\
\hline\hline
\thead{ImVoxel\\Net [29]}&\thead{WACV\\2022}&\thead{Mono.+\\cam. pose}&\thead{2D+\\3D}&45.12&79.17&13.27&63.07&31.20&{\bf 35.45}&38.38&60.59&{\bf 51.14}&19.24&43.66\\
\thead{T3DU[30]}&\thead{CVPR\\2020}&\thead{Mono.+\\geo.}&\thead{2D+\\3D}&18.05&60.65&5.04&17.55&27.93&21.19&17.01&44.90&36.48&14.51&26.38\\
\thead{IM3D[31]}&\thead{CVPR\\2021}&\thead{Mono.+\\extra data}&\thead{2D+\\3D}&33.81&{\bf 89.32}&11.90&35.14&{\bf 49.03}&29.27&41.34&{\bf 69.10}&57.37&{\bf 33.93}&45.21*\\
\thead{Perspecti\\veNet[28]}&\thead{NeurIPS\\2019}&\thead{Mono.}&\thead{2D+\\3D}&41.35&79.69&13.14&40.42&20.19&/&/&62.35&44.12&/&/\\
\hline
Ours&/&\thead{Mono.+\\pseudo-\\view}&\thead{2D}&{\bf 65.44}&76.32&{\bf 22.48}&{\bf 70.66}&18.06&32.02&{\bf 56.19}&58.71&42.85&6.80&{\bf 44.95}\\
\hline
\end{tabular}
\end{center}

\end{table*}

\subsection{Data Efficiency}
To demonstrate the efficiency of our method in terms of data utilization, we conduct experiments on the KITTI dataset. As the Pedestrian (4,487 samples) and Cyclist (1,627 samples) categories in the KITTI dataset are significantly smaller in size than the Car category (28,742 samples), they are chosen as benchmarks for this experiment. This selection is made to effectively evaluate the ability of our method to handle smaller datasets. The proposed method is trained using varying proportions of the available training data, specifically 80\%, 50\%, and 25\% respectively. The performance of the trained models is subsequently evaluated on the KITTI test set. Tab. III reports the 3D detection performance of Pedestrian and Cyclist on the KITTI test set. Our method greatly outperforms previous state-of-the-art monocular-based methods on all evaluated categories, even when utilizing only 25\% of the training data. 

\begin{table*}
\begin{center}
\caption{{\bf 3D Detection performance of Pedestrian and Cyclist on the KITTI test set.  } {\bf Bold} is used to highlight the best results.}
\renewcommand{\arraystretch}{1.4}
\begin{tabular}{l| p{0.75cm} p{0.45cm} c c c c c}
\hline
\multirow{2}{*}{Method} & \multirow{2}{*}{Ref.} & \multirow{2}{*}{Data} & \multirow{2}{*}{Input} & \multicolumn{2}{c}{AP$_{R40}$ [Easy / Mod / Hard]} & \multicolumn{2}{c}{AP$_{R40}$ [Easy / Mod / Hard]}\\
\cline{5-8}
& & & &\small AP$_{3D}@IoU=0.5$ & \small AP$_{BEV}@IoU=0.5$ &\small AP$_{3D}@IoU=0.5$ &\small AP$_{BEV}@IoU=0.5$ \\
\hline\hline
\thead{LPCG-\\Monoflex[40]}&{\small \thead{ECCV\\2022}}&100\%&\thead{Image+\\LiDAR}&10.82 / 7.33 / 6.16&12.11 / 7.92 / 6.61&6.98 / 4.38 / 3.56&8.14 / 4.90 / 3.86\\
\thead{DEVI\\ANT[39]}&{\small \thead{ECCV\\2022}}&100\%&\thead{Image+\\depth}&13.43 / 8.65 / 7.69&14.49 / 9.77 / 8.28&5.05 /3.13 / 2.59&6.42 / 3.97 / 3.51\\
\thead{DD3D[38]}&{\small \thead{ICCV\\2021}}&100\%&\thead{Image+\\depth}&13.91 / 9.30 / 8.05&15.90 / 10.85 / 8.05&7.52 / 4.79 / 4.22&9.20 / 5.69 / 5.20\\
\thead{PS-fld[37]}&{\small \thead{CVPR\\2022}}&100\%&\thead{Image+\\LiDAR}&16.95 / 10.82 / 9.26&19.03 / 12.23 / 10.53&11.22 / 6.18 / 5.21&12.80 / 7.29 / 6.05\\
\thead{OPA-3D[35]}&{\small \thead{R-AL\\2022}}&100\%&\thead{Image+\\LiDAR}&15.65 / 10.49 / 8.80&17.14 / 11.01 / 9.94&5.16 / 3.45 / 2.86&6.01 / 3.75 / 3.56\\
\thead{Mono\\DTR[36]}&{\small \thead{CVPR\\2022}}&100\%&\thead{Image+\\LiDAR}&15.33 / 10.18 / 8.61&16.66 / 10.59 / 9.00&5.05 / 3.27 / 3.19&5.84 / 4.11 / 3.48\\
\hline

\multirow{3}{*}{Ours}&\multirow{3}{*}{/}&80\%&\multirow{3}{*}{\thead{Image+\\pseudo\\-view}}&40.19 /{\bf 31.89}/{\bf 28.32}&{\bf 52.26}/42.93/37.34&{\bf20.02 }/{\bf 13.93}/{\bf 12.48} & {\bf 28.51}/{\bf 21.82}/{\bf 18.94}\\
&&50\%&&{\bf 40.85}/31.60/27.96&51.14/{\bf 44.11}/{\bf 38.39}&16.66/13.17/11.18&21.70/17.70/15.28\\
&&25\%&&37.50/30.24/26.72&50.08/43.52/38.03&13.69/11.22/9.45&19.55/15.80/13.60\\

\hline
\end{tabular}
\end{center}

\end{table*}

\subsection{3D Annotator Using RCV}
To justify that our method does not consume any 3D annotations and can be viewed as an automatic 3D annotator, we formulate two annotated datasets, named “3D\_HUMAN” and “3D\_HAND”, using RCV. All data is collected by an Azure Kinect DK. Following the 2D annotating method in Fig. 6, we label 1,600 2D bounding boxes for “3D\_HUMAN” and 530 2D bounding boxes for “3D\_HAND”. Note that these 2D bounding boxes are labeled on ‘pseudo-view’ images. Annotations for the original image are not mentioned as they are quite straightforward. After training, we obtain two RCV models, which can produce 3D bounding boxes for humans and hands respectively. “3D\_HUMAN” contains fully annotated humans in about 30 indoor scenes, as shown in Fig. 7. It consists of approximately 1,500 frames of data, each of which includes a RGB image, a point cloud, and one or more 3D bounding boxes. Totally, it contains more than 4,500 3D bounding boxes generated by RCV.

\begin{figure*}

\begin{center}
   \includegraphics[width=0.8\linewidth]{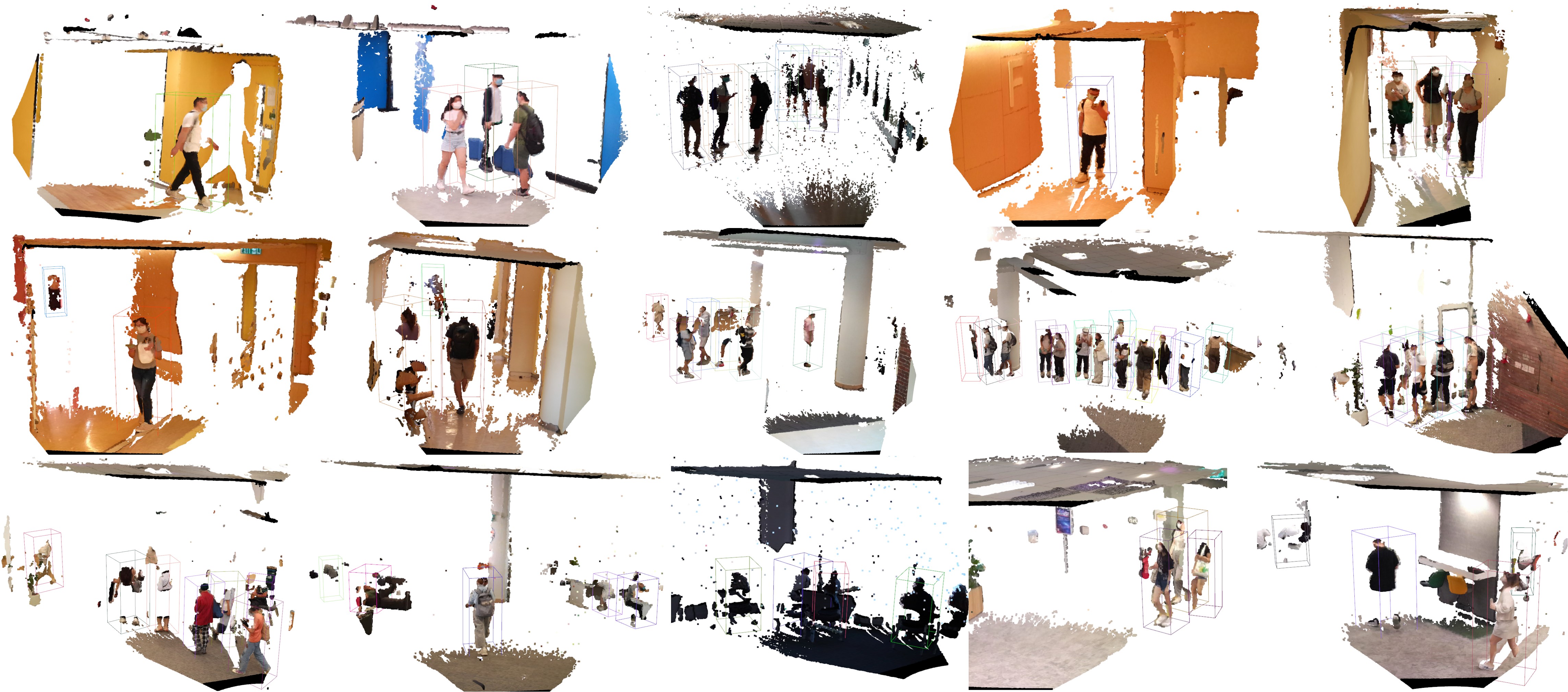}
\end{center}
   \caption{{\bf 3D boxes generated by RCV on '3D\_HUMAN'.}}
\label{fig:long}
\label{fig:onecol}
\end{figure*}

“3D\_HAND” contains fully annotated hands from 8 participants. It consists of 1500 frames of data. Totally, this dataset contains about 1,500 fully oriented 3D bounding boxes generated by RCV, as shown in Fig. 8. We argue that the final datasets can be used to at least pretrain some 3D detection models after slight manual selection and adjustment. Therefore, we believe that it is feasible to use RCV as a preliminary 3D annotation tool. In the future, we will train some 3D detectors on our datasets. Similarly, if one wants to realize any 3D object detection in different scenarios, the same steps can be performed relying on RCV. We just need to collect the data and label some 2D images, which is very simple compared to 3D labeling.

\begin{figure*}

\begin{center}
   \includegraphics[width=0.85\linewidth]{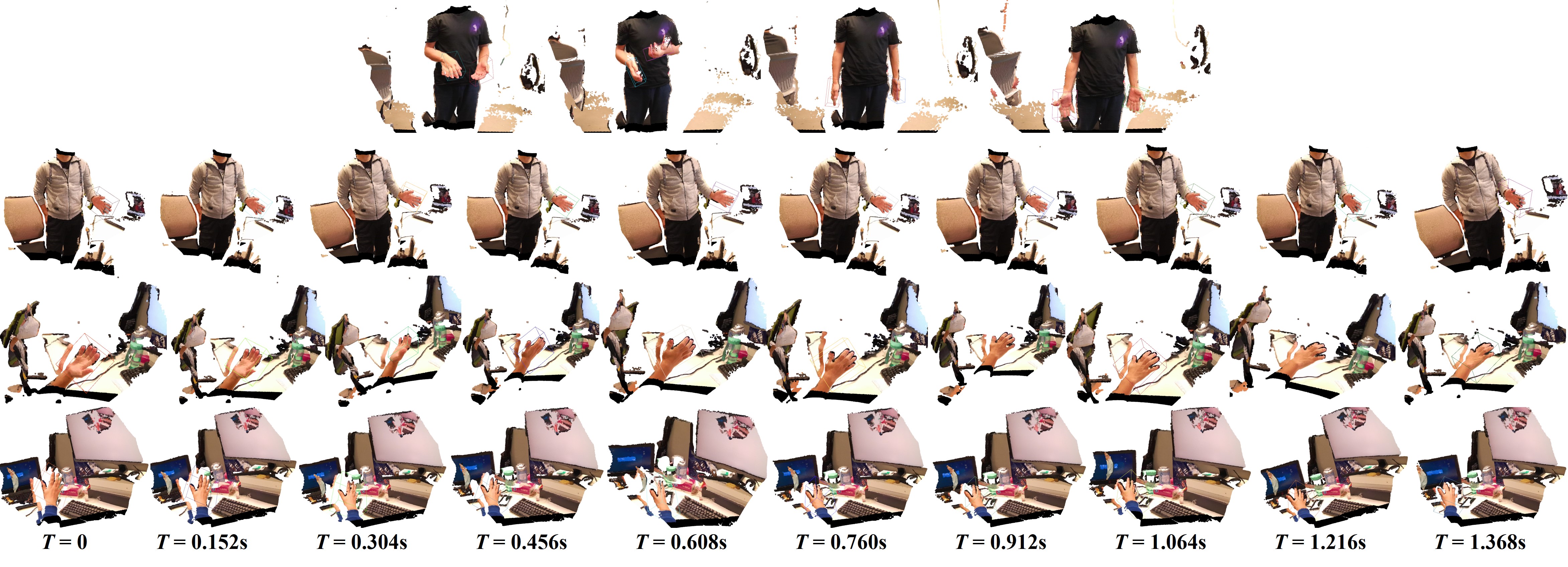}
   \captionsetup{justification=centering}
\end{center}
   \caption{{\bf 3D boxes generated by RCV on '3D\_HAND'.}}
\label{fig:long}
\label{fig:onecol}
\end{figure*}

\subsection{3D Detection on A Depth Camera}
To justify that our method can detect in real-time in the real world, we apply RCV on an Azure Kinect DK. Specifically, we leverage the hand detection model described in Section III.D to detect a hand and generate a fully oriented 3D bounding box. The system achieves a frequency of approximately 7Hz. The full video and the code can be accessed in the supplementary materials.

\section{Discussions and Future Works}
\subsection{Imitate 3D Labeling Process}
In fact, our method imitates the manual 3D labelling process [1], where the annotator first places a coarse bounding box and then rotates the box multiple times while manually detecting 2D bounding boxes on three different views to obtain a 3D annotation. Our method imitates this process by replacing ’rotate the box’ with generating projection axes (as described in Section II.D), ’manually detects 2D bounding boxes’ with YOLOv5, and ’multiple times’ with a recursive process. Finally, our method can perform this 3D annotation process automatically.

\subsection{Automatic Labeling Pipeline and Datasets}
Our method can achieve (semi-) automatic 3D annotations with some 2D bounding boxes from scratch. This is a significant practical contribution to scenarios without any annotated data. Compared to some existed semi-automatic labeling pipeline, for example H2O [43] that leveraged already trained DenseFusion [44] to label the data, our pipeline has a distinct practical advantage. The two obtained datasets can be used to train many existed 3D detectors, which will be our future work. Notably, 3D\_HAND comprises fully-oriented 3D box annotations.

\subsection{Limitations}
In some cases, the method may fail to converge, resulting in failure to detect the object: (1) Poor performance of the trained 2D detector, which filters out points belonging to the object in each iteration, making it difficult for the system to converge, and (2) the 2D detector fails to detect the object in the projected images, resulting in early stopping of the detection process. We found in our experiments that the first case rarely occurs. Instead, the more common scenario is the second case. Our method heavily relies on the performance of the 2D detector.

Furthermore, we observe that our method performs relatively poorly on larger-sized objects. An intuitive explanation is that larger objects are more likely to have a smaller proportion of their regions captured by the camera, potentially resulting in decreased performance. However, it needs to be experimentally verified in the future work.

\section{Conclusion}
We propose a new 3D detection method, named RCV, that does not consume 3D labels and yields fully oriented 3D boxes on point clouds. In the future, we will explore the application of this method in scenarios such as human-following robots and robotic grasping tasks. Currently, we have only explored one type of divide-and-conquer recursive strategy, and there is still a need for further research on additional recursive strategies.



\end{document}